\documentclass[sigconf]{acmart}

\usepackage{tikz}
\usepackage{multirow}
\usetikzlibrary{positioning, shapes.geometric, arrows.meta, fit}

\AtBeginDocument{%
  }

\copyrightyear{2026}
\acmYear{2026}
\setcopyright{cc}
\setcctype{by}
\acmConference[SIGIR '26]{Proceedings of the 49th International ACM SIGIR Conference on Research and Development in Information Retrieval}{July 20--24, 2026}{Melbourne, VIC, Australia}
\acmBooktitle{Proceedings of the 49th International ACM SIGIR Conference on Research and Development in Information Retrieval (SIGIR '26), July 20--24, 2026, Melbourne, VIC, Australia}
\acmDOI{10.1145/3805712.3808414}
\acmISBN{979-8-4007-2599-9/2026/07}

\begin{document}

\title{Surface-Form Neural Sparse Retrieval: Robust Fuzzy Matching for Industrial Music Search}

\author{Paul Greyson}
\affiliation{%
  \institution{Amazon}
  \city{San Francisco}
  \state{CA}
  \country{USA}
}
\email{pgreyson@amazon.com}

\author{Zhichao Geng}
\affiliation{
  \institution{Amazon}
  \city{Shanghai}
  \country{China}
}
\email{zhichaog@amazon.com}

\author{Wei Zhang}
\affiliation{%
  \institution{Amazon}
  \city{San Francisco}
  \state{CA}
  \country{USA}
}
\email{wizhang@amazon.com}

\author{Yang Yang}
\affiliation{%
  \institution{Amazon}
  \city{Shanghai}
  \country{China}
}
\email{yych@amazon.com}

\renewcommand{\shortauthors}{Paul Greyson, Zhichao Geng, Wei Zhang, and Yang Yang}

\begin{abstract}
Music search at the scale of Amazon Music presents a unique challenge: queries frequently deviate from indexed metadata due to misspellings, transpositions, and phonetic variations, yet the retrieval system must operate under strict millisecond-level latency constraints. 
Our existing learning-to-retrieve system, the High Confidence Index (HCI), learns query-entity associations from customer behavior, relying on continual ``exploration'' to choose candidates. Traditional n-gram matching enables this exploration but suffers from poor semantic robustness and high noise, limiting the system's ability to learn from long-tail queries.
In this work, we present a \textbf{robust neural sparse retrieval system} designed to maximize exploration efficiency. We adapt a state-of-the-art \textbf{inference-free} sparse retrieval architecture to the music domain, combining it with an effective \textbf{domain-specific granular subword tokenization strategy}. Our approach utilizes short-length token constraints (max 3 chars) to enforce the learning of surface-form robustness over lexical memorization.
By pre-computing the neural embeddings and term expansions during the offline indexing phase, online processing is reduced to minimal tokenization and IDF weighting, achieving effectively zero latency overhead for query encoding.
Evaluations on a 6M-document production corpus show an aggregate \textbf{91.4\%} recall@10 (vs. \textbf{57.7\%} for trigrams) at comparable throughput. Simulation of the HCI feedback loop demonstrates improved exploration efficiency, with \textbf{+0.8\%} higher stabilized recall than production trigrams. Ablation studies indicate that our sparse training methodology drives the performance gains, while domain-specific pretraining provides a cost-effective alternative to large-scale general-purpose pretraining.
\end{abstract}



\maketitle
\section{Introduction}

Music search queries frequently deviate from indexed metadata due to ``query misspecifications'' ranging from phonetic misspellings (``tayler'' vs. ``taylor'') to character variations and transpositions~\cite{brill2000improved, martins2004spelling}. As illustrated in \textbf{Table~\ref{tab:query_challenges}}, these deviations present a dual challenge: the system must be robust enough to bridge significant lexical gaps yet efficient enough to meet strict millisecond-level latency budgets.

\begin{table}[t]
\centering
\caption{Common cases of query misspecifications in music search.}
\label{tab:query_challenges}
\resizebox{\columnwidth}{!}{
\begin{tabular}{l p{3.5cm} l}
\toprule
\textbf{Category} & \textbf{User Query} & \textbf{Target Entity} \\
\midrule
Simple Misspelling & tayler swift & Taylor Swift \\
Character Variation & p!nk & Pink \\
Transposition & carpenter sabrina & Sabrina Carpenter \\
Incidental Terms & taylor swift songs & Taylor Swift \\
\bottomrule
\end{tabular}
}
\end{table}

\usetikzlibrary{shapes, arrows.meta, positioning, fit, backgrounds, shadows.blur, calc}

\definecolor{techblue}{RGB}{232, 241, 250}
\definecolor{techborder}{RGB}{70, 130, 180}
\definecolor{processgreen}{RGB}{235, 247, 235}
\definecolor{processborder}{RGB}{34, 139, 34}
\definecolor{alertred}{RGB}{255, 235, 235}
\definecolor{alertborder}{RGB}{205, 92, 92}
\definecolor{coregray}{RGB}{245, 245, 245}

\begin{figure*}[t]
\centering
\begin{tikzpicture}[
  node distance=1.2cm and 1.5cm,
  font=\sffamily\footnotesize,
  base/.style={
    rectangle, 
    rounded corners=3pt, 
    draw=gray!40, 
    thick, 
    minimum height=1cm, 
    text centered, 
    blur shadow={shadow blur steps=5}
  },
  query/.style={base, fill=techblue, draw=techborder, text width=2.2cm},
  action/.style={base, fill=processgreen, draw=processborder, text width=2.8cm},
  result/.style={base, fill=alertred, draw=alertborder, text width=1.8cm},
  feedback/.style={base, fill=orange!10, draw=orange!60, text width=2cm},
  entry/.style={
    rectangle, 
    draw=gray!30, 
    fill=white, 
    text width=2.8cm, 
    minimum height=0.6cm, 
    font=\scriptsize\ttfamily,
    inner sep=3pt
  },
  line/.style={->, >=latex, thick, color=gray!70},
  feedbackline/.style={->, >=latex, thick, color=orange!80, dashed}
]

  \node[query] (search) {\textbf{Search Query}\\``tayler swift songs''};

  \node[entry, right=0.6cm of search, yshift=0.4cm] (old_entry) {``taylor swift''\\$\to$ Taylor Swift};
  \node[entry, below=0.1cm of old_entry, fill=gray!10, dashed] (new_entry) {``tayler swift songs''\\$\to$ Taylor Swift};
  
  \node[above=0.6cm of old_entry, font=\bfseries\small, color=gray!70] (index_label) {HCI Index};

  \node[action, right=1.0cm of old_entry, yshift=-0.4cm] (match) {\textbf{Neural Sparse Fuzzy Match}\\``tayler...'' $\approx$ ``taylor...''};

  \node[result, right=1.0cm of match] (result) {\textbf{Result}\\Taylor Swift};

  \node[feedback, below=0.7cm of match] (customer) {\textbf{Customer Engagement}\\(Click/Play)};

  \begin{scope}[on background layer]
    \node[fit=(old_entry)(new_entry)(index_label), 
          draw=gray!30, fill=coregray, rounded corners, inner sep=5pt,
          label={[anchor=south, font=\tiny, color=gray!50]south:Storage}] (index_box) {};
  \end{scope}


  \draw[line] (search) -- (index_box.west |- search);

  \draw[line] (old_entry.east) -- (match.west |- old_entry.east);

  \draw[line] (match) -- (result);

  \draw[feedbackline] (result.south) |- (customer.east);
  \draw[feedbackline] (customer.west) -| node[pos=0.2, above, font=\tiny\sffamily, color=orange!80] {Learns Association} (new_entry.south);

\end{tikzpicture}
\caption{HCI exploration feedback loop. \textbf{Neural sparse fuzzy matching} (green) enables the system to bridge the gap between misspelled queries and canonical entities. Successful customer engagement (orange) creates a permanent record in the index (dashed box), converting future fuzzy matches into exact matches.}
\label{fig:hci}
\end{figure*}
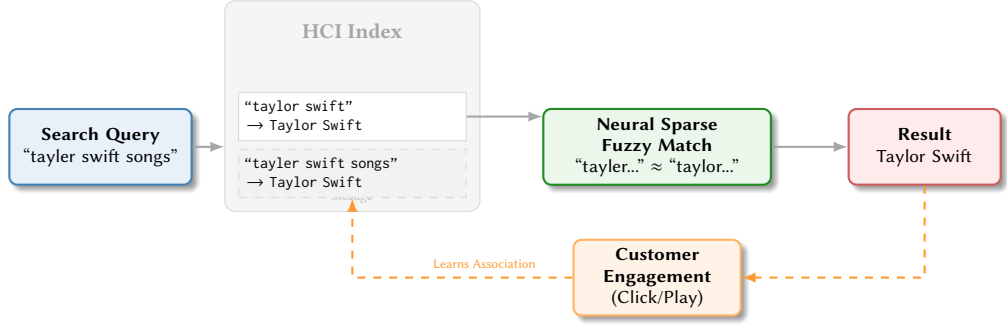

One method that Amazon Music uses to address relevance is the High Confidence Index (HCI), a learning-to-retrieve system that effectively memorizes query-entity affinities from historical engagement. However, HCI's learning depends critically on \textit{exploration}—successfully matching a query to a candidate for the first time. Currently, exploration relies on traditional trigram matching. These methods are semantically brittle and technically constrained. For instance, OpenSearch's trigram configurations enforce minimum token lengths, discarding critical short terms such as "me". On the other hand, avoiding this loss by including whitespace tokens introduces order sensitivity that hinders matching transposed queries. As illustrated in Table~\ref{tab:examples}, they often fail to distinguish incidental terms such as ``songs'' in ``taylor swift songs'' or break under structural variations like ``p!nk'', returning only exact matches or unrelated extensions. These limitations hinder the system's ability to efficiently explore and learn from long-tail queries.

\textbf{Learned Sparse Retrieval (LSR)} has emerged as a compelling solution to these challenges~\cite{formal2021splade, dai2020context}. By predicting sparse weight distributions over a vocabulary, LSR combines the contextual surface-form understanding of neural models with the efficiency of exact matching. Crucially, unlike dense retrieval~\cite{dpr} which requires specialized vector databases (ANN)~\cite{malkov2018efficient, johnson2019billion} and adds latency from online inference, LSR retains compatibility with standard \textbf{inverted indices}~\cite{bialecki2012lucene4, yang2017anserini}. This ensures high cost-effectiveness, seamless integration with existing production infrastructure, and inherent interpretability of matching features.

Building on this paradigm, we present a \textbf{robust neural sparse retrieval system} optimized for strict industrial latency. While standard LSR models (e.g., SPLADE~\cite{formal2021splade}) often require costly online inference, we adopt an \textbf{inference-free asymmetric architecture}~\cite{splade2, geng2024towards, shen2025exploring}. By pre-computing document representations via neural expansion during offline indexing, and using lightweight IDF weighting for online queries, we maintain the benefits of LSR with effectively zero latency overhead for query encoding.

To further address query misspecifications, we adopt a \textbf{data-driven granular tokenization strategy}. Instead of relying on rigid n-grams, our approach learns granular sub-word patterns~\cite{bojanowski2017fasttext, kudo2018sentencepiece, sennrich2016bpe} directly from search logs~\cite{radlinski2007active, joachims2002optimizing}, enabling the model to capture surface-form robustness. This allows the system to recognize the structural equivalence between variations without exact lexical matching. As demonstrated in the neural sparse column of Table~\ref{tab:examples}, this strategy robustly handles complex variations like ``p!nk'' and ``tayler swift'' that typically fail with traditional methods.

\begin{table}[t]
\caption{Comparison of neural sparse vs trigram matching on music queries. Scores shown in parentheses.}
\label{tab:examples}
\resizebox{\columnwidth}{!}{
\begin{tabular}{p{1.5cm}p{3.5cm}p{3.5cm}}
\toprule
\textbf{Query} & \textbf{Neural Sparse} & \textbf{Trigram} \\
\midrule
p!nk & p!nk (22.4), p nk (20.1), pnk (18.3) & p!nk (398), million dreams p!nk (28.9) \\
\midrule
tayler swift & tayler swift (24.6), taylor swift (19.9), tyler swift (19.3) & tayler swift (458), tayler swift \& post malone (89.1) \\
\bottomrule
\end{tabular}
}
\end{table}

Our main contributions are:
\begin{itemize}
    \item The successful adaptation and deployment of an inference-free sparse retrieval architecture (SPLADE) to industrial music search, requiring no online neural encoding.
    \item An empirical validation of a domain-specific, granular tokenization strategy that outperforms traditional trigrams in robust fuzzy matching.
    \item Offline multilingual evaluations on a 6M-document production corpus showing \textbf{91.4\%} aggregate recall@10 (vs. \textbf{57.7\%} for trigrams) at comparable throughput, effectively achieving zero latency overhead for query encoding, alongside simulations demonstrating improved exploration efficiency.
    \item Ablation studies revealing that our sparse training pipeline is the primary performance driver, and that a modest domain-specific pretraining effectively replaces large-scale general-purpose pretraining.
\end{itemize}

\section{System Overview}

A key mechanism for fuzzy music search is the High Confidence Index (HCI), a probabilistic retrieval framework that ranks entities based on historical query-entity affinity. The system estimates the relevance score between a user query $Q$ and an entity $E$ through a factorization over previously observed queries $Q'$:

\begin{equation}
\label{eq:hci}
\text{Score}(Q, E) \approx \sum_{Q' \in \mathcal{I}} \underbrace{P(Q'|Q)}_{\text{Query Sim.}} \cdot \underbrace{P(E|Q')}_{\text{Behavioral}}
\end{equation}

Here, $P(E|Q')$ represents the historical engagement probability stored in the index. Our work targets the \textbf{Query Similarity} term $P(Q'|Q)$. Although HCI effectively serves frequent queries via exact matching, it relies on an \textit{exploration} mechanism to handle novel or misspecified queries (e.g., $Q=$``tayler'', $Q'=$``taylor'').

As illustrated in Figure~\ref{fig:hci}, the system operates as a continuous feedback loop. Our \textbf{Neural Sparse Retrieval} module acts as the exploration engine, designed to robustly estimate $P(Q'|Q)$. To meet strict latency budgets, we use an \textbf{asymmetric inference-free architecture}~\cite{geng2024towards}:

\begin{itemize}
    \item \textbf{Offline Document Encoding:} In our retrieval framework, the "documents" to be indexed are historical queries ($Q'$). These are processed offline by our sparse encoding model, where terms are expanded and sparse weights are predicted and stored in the OpenSearch inverted index.
    \item \textbf{Online Zero-Overhead Query Encoding:} At query time, we bypass neural inference. The user query $Q$ is processed solely by our granular subword tokenizer and weighted via IDF.
\end{itemize}

Implementation-wise, this retriever operates as a pluggable module within the OpenSearch query plan. It replaces the legacy trigram matcher to provide a more robust similarity signal. When a user engages with a candidate retrieved via this signal, the system captures the event and updates the Behavioral term $P(E|Q')$, effectively converting a fuzzy exploration problem into a precise memory retrieval for future users.

\section{Methodology}

We adapt and optimize a neural sparse retrieval framework specifically for the lexical irregularities of music search. Our approach diverges from standard SPLADE LSR model~\cite{formal2021splade, splade2, formal2022distillation, lassance2024splade} by enforcing strict sub-lexical constraints and adopting a fully inference-free query path~\cite{geng2024towards, shen2025exploring, nardini2025effective}.

\subsection{Surface-Form Robust Pre-training}
\label{sec:pretraining}
Standard PLMs typically rely on fixed subword vocabularies (e.g., WordPiece), where minor perturbations can induce sharply different tokenizations and degrade robustness to severe query misspecifications in search \cite{schuster2012japanese,el-boukkouri-etal-2020-characterbert}.
For example, a typo like ``tayler'' can map to a substantially different subword sequence than ``taylor'', reducing shared lexical evidence for the model to learn their equivalence.
We address this via a joint tokenization and pre-training strategy tailored to music query noise. It is important to note that our design targets \textit{robust lexical matching} rather than broad semantic retrieval. By constraining tokens to short lengths, we focus the model on learning surface-form equivalence (e.g., ``ph'' $\approx$ ``f'') rather than semantic synonyms, which minimizes the risk of irrelevant ``hallucinations'' common in dense retrieval.

\textbf{Granular Tokenization.} 
We train a custom SentencePiece Unigram model~\cite{kudo2018sentencepiece, kudo2018subword} on 20 million music queries with a strict constraint: \textbf{maximum token length of 3 characters}. 
This constraint acts as an effective empirical heuristic, encouraging the tokenizer to decompose terms into granular subword constituents (e.g., character n-grams) rather than memorizing full words. Consequently, variations like ``p!nk'' and ``pink'' share significant sub-token overlap, providing a robust initialization for retrieval.

\textbf{Domain-Specific MLM.} Using this granular vocabulary ($|V|=32,000$), we pre-train a BERT-base encoder from scratch on the query corpus. We employ the Masked Language Modeling (MLM) objective to teach the model the contextual semantics of these short sub-lexical units~\cite{devlin-etal-2019-bert}. This produces a foundation model that is natively aligned with the structural irregularities of music search queries, unlike generic BERT models fine-tuned on standard text~\cite{beltagy2019scibert, gururangan2020don}.

\subsection{Asymmetric Sparse Architecture}
We adopt an inference-free architecture~\cite{splade2}, utilizing a SPLADE-doc encoder to decouple neural computation from online retrieval.

\textbf{Offline Document Encoding.} 
Historical queries ($Q'$), treated as documents $d$, are encoded using the robust BERT model. SPLADE-doc repurposes the MLM head to project token embeddings $\mathbf{h}_t$ to the vocabulary space. The weight $w_j^{(d)}$ for token $j$ is derived via max-pooling:
\begin{equation}
w_j^{(d)} = \max_{t \in d} \left( \log(1 + \text{ReLU}(\mathbf{E}_{j}^T \mathbf{h}_t + b_j)) \right)
\end{equation}
This \textit{neural expansion} assigns weights to unobserved but related tokens (e.g., misspellings). Optimized with InfoNCE~\cite{dpr} and FLOPS regularization~\cite{paria2020minimizing}, it yields a sparse, robust representation compatible with inverted indices.

\textbf{Online Query Weighting.} 
To ensure zero-latency inference, we bypass the neural encoder at query time. Query weights $w_j^{(q)}$ are derived solely from tokenization and corpus statistics~\cite{geng2024towards}:
\begin{equation}
w_j^{(q)} = \mathbb{I}(v_j \in q) \cdot \text{IDF}(v_j)
\end{equation}
This limits online computation to tokenization while retaining the expressiveness of offline expansions.

\textbf{Indexing and Scoring.} We implement dot-product retrieval $\sum w_j^{(q)} w_j^{(d)}$ via OpenSearch's \texttt{rank\_features} field, which stores sparse weights as 16-bit floats for high-performance exact scoring.

\textbf{Optimization Objective.} We fine-tune the encoder using InfoNCE loss with hard and in-batch negatives, alongside FLOPS regularization to enforce sparsity:
\begin{equation}
\mathcal{L} = \mathcal{L}_{\text{InfoNCE}} + \lambda_{reg} \cdot \sum_{j \in \mathcal{V}} \left( \frac{1}{N} \sum_{i=1}^{N} w_j^{(d_i)} \right)^2
\label{eq:FLOPS_loss}
\end{equation}
where $N$ is the batch size and $d_i$ is the $i$-th document.

\subsection{Weakly Supervised Data Construction}

Since manual relevance labeling is infeasible at scale, we construct a weakly supervised dataset from aggregated behavioral logs.

\textbf{Data Mining Strategy.} We utilize 28 days of high-confidence query-entity playback logs. To train the model specifically for robust fuzzy matching (rather than broad semantic matching), we construct positive pairs $(q, q^+)$ by aligning queries that:
(1) Led to a playback of the \textit{same} entity,
(2) Have a length ratio $\text{len}(q_{\text{short}}) / \text{len}(q_{\text{long}}) \ge 0.8$, and
(3) Possess a character-level Levenshtein distance within a length-scaled threshold: $\max(1, \lfloor\text{len}(q) / 10\rfloor)$.
This serves as an effective heuristic to capture common misspellings (``tayler swift'' $\leftrightarrow$ ``taylor swift''), spacing differences (``sonideroaczino'' $\leftrightarrow$ ``sonidero aczino''), transliterations (``radha kawach'' $\leftrightarrow$ ``radha kavach''), and other lexical variations.

\textbf{Hard Negatives.} Generated through iterative mining with sparse retrieval. Epoch 1 uses sparse scores from the base BERT initialization via the MLM head; subsequent epochs use the previous sparse checkpoint for mining. We filter pairs sharing any playback entity to avoid false negatives.

\textbf{Data Splitting.} To prevent data leakage, we partition the dataset using connected components on the query-entity bipartite graph. This ensures that no query or entity in the test set has been seen during training, strictly evaluating the model's generalization capability. Initial empirical observations indicated that random splitting led to severe data leakage and artificially inflated performance, making this graph-based partitioning a critical best practice for evaluation.

\section{Experiments}

\subsection{Setup}
\textbf{Data \& Implementation.} We curate 1.19M documents and 237K test queries, utilizing graph-based partitioning to prevent leakage. Our best model is trained from scratch for $\sim$16 hours on a p4d.24xlarge instance ($8\times$ A100 GPUs), running to 1,040,000 steps. We use an effective batch size of 120, a learning rate of $8.5 \times 10^{-6}$ (2,000-step warmup), and progressive hard negative mining (4 negatives per query).

\textbf{Metrics.} We report Recall (R@k), Precision (P@k), and NDCG (N@k) for relevance. Efficiency is measured by query latency (ms) and index storage. Exploration capability is evaluated via simulation Recall against behavioral ground truth.

\subsection{Offline Evaluation}

\textbf{Retrieval Performance.} Table~\ref{tab:results} compares our system against the production HCI Trigram baseline and fuzzy edit-distance on the production HC corpus (6.04M documents, 237K multilingual test queries). Our domain-trained neural sparse model reaches \textbf{91.4\%} aggregate recall@10 compared to 57.7\% for HCI Trigram and 83.4\% for fuzzy edit-distance. All methods are measured on the same OpenSearch cluster (p4d.24xlarge, concurrency 32); neural sparse throughput is comparable to trigram matching (2,471 vs. 2,299 QPS) while producing substantially richer document representations. This confirms that surface-form robust modeling effectively bridges the lexical gaps in misspecified queries across diverse languages where rigid string matching cannot, while the inference-free query path preserves production throughput.

\begin{table}[h]
\centering
\caption{Retrieval performance. Neural Sparse consistently outperforms the production HCI Trigram baseline on the production HC corpus (6.04M docs, 237K test queries, 8 languages). QPS on p4d.24xlarge, concurrency 32.}
\label{tab:results}
\small
\resizebox{0.8\columnwidth}{!}{
\begin{tabular}{lccc}
\toprule
\textbf{Method} & \textbf{N@1} & \textbf{R@10} & \textbf{QPS} \\
\midrule
Neural Sparse (ours) & \textbf{.557} & \textbf{.914} & 2471 \\
Fuzzy Edit-Distance & .508 & .834 & 1465 \\
HCI Trigram (prod.) & .475 & .577 & 2299 \\
Vanilla Trigram & .316 & .521 & 2634 \\
\bottomrule
\end{tabular}
}
\end{table}

\textbf{Efficiency \& Cost.} 
Latencies are minimized via the inference-free design: query encoding overhead is effectively zero (tokenization only), comparable to BM25. Sparse vectors add only $\sim$25 tokens per document, increasing index size by a modest $\sim$30\%. 

\subsection{Ablation Studies}

To isolate the contributions of our three design choices---domain pretraining, the sparse training pipeline, and the granular tokenizer---we evaluate three ablations on the same HC corpus (see Table~\ref{tab:ablation}).

\textbf{A. Off-the-shelf sparse retriever.} Evaluating an OpenSearch-released multilingual sparse model~\cite{geng2024towards} without domain training achieves only R@10=0.795, underperforming fuzzy edit-distance (0.834). This confirms that general-purpose sparse models require domain adaptation for character-level lexical matching.

\textbf{B. Domain data with a general-purpose backbone.} Applying our full sparse training pipeline to a multilingual BERT backbone (without our domain pretraining) achieves R@10 between 0.906 and 0.937, depending on FLOPS regularization. This shows that our sparse training pipeline is the primary driver of quality, accounting for most of the gain over the off-the-shelf baseline.

\textbf{C. Domain data with standard WordPiece tokenization.} Substituting our granular SentencePiece tokenizer with a standard 24K WordPiece tokenizer (keeping domain pretraining and sparse pipeline) yields R@10=0.909 vs 0.949 for our SentencePiece tokenizer on a smaller evaluation corpus. By preventing canonical forms and misspellings from decomposing into entirely different sub-tokens, our 3-character ceiling constraint improves robustness to lexical perturbations.

\textbf{Summary.} The sparse training methodology is the primary performance driver. However, our domain BERT pretrained on just 20M queries matches or beats a multilingual BERT pretrained on billions of tokens. For teams with limited computational resources, this modest in-domain corpus provides a highly cost-effective substitute for large-scale general-purpose pretraining.

\begin{table}[h]
\centering
\caption{Ablation summary on the HC corpus (6.04M docs). ``Dims'' denotes average non-zero dimensions per document.}
\label{tab:ablation}
\small
\resizebox{\columnwidth}{!}{
\begin{tabular}{llccc}
\toprule
\textbf{Model} & \textbf{Tokenizer / Backbone} & \textbf{R@10} & \textbf{Dims} & \textbf{QPS} \\
\midrule
Ours (end-to-end) & SentencePiece-3 / Domain BERT & .914 & 86 & 2471 \\
Domain WordPiece & WordPiece-24K / Domain BERT & .909\textsuperscript{\textdagger} & --- & --- \\
BERT-ml (T1) & WordPiece-119K / BERT-ml & .906 & 37 & 3953 \\
BERT-ml (T9) & WordPiece-119K / BERT-ml & \textbf{.937} & 102 & 2026 \\
Off-the-shelf AOS & WordPiece-105K / BERT-ml distill & .795 & 206 & 2152 \\
Fuzzy Edit-Distance & --- & .834 & --- & 1465 \\
HCI Trigram (prod.) & --- & .577 & --- & 2299 \\
\bottomrule
\multicolumn{5}{l}{\scriptsize \textsuperscript{\textdagger} Evaluated on a separate smaller corpus.}
\end{tabular}
}
\end{table}

\subsection{Production Log Replay}

To assess the system's long-term evolution in a realistic setting, we conducted a replay backtest using production search logs. This evaluation simulates the continuous HCI feedback loop, where retrieved entities validated by user behavior are subsequently learned (indexed) for future retrieval.

We used 7 days of high-confidence behavioral logs (queries with $>$3 clicks) from production traffic. The simulation operates in discrete epochs, where each epoch represents one day of system operation. At epoch 0 (cold start), the index contains only catalog metadata with no behavioral data, and retrieval relies solely on standard text matching against entity fields (title, artist name, etc.). In subsequent epochs, we perform a partial index update: for each query in the evaluation set, we retrieve candidates using the current retrieval configuration, and successfully retrieved entities that match the behavioral ground truth are written back to the index as query-entity pairs with their engagement scores for use in the next epoch. We evaluate against $\sim$750K queries from the 7-day log period where entities received customer clicks, representing validated relevance judgments from real user behavior.

We compare three retrieval strategies: (1) Cold Start using only text matching without behavioral data, (2) production baseline using exact match on historical queries plus trigram-based fuzzy matching, and (3) our approach using exact match plus Neural Sparse retrieval (limited to top-10 candidates to reduce noise from lower-confidence matches).

Table~\ref{tab:simulation} shows that Neural Sparse significantly improves the discovery process. We ran the simulation for 15 epochs, with both methods converging by epoch 11. Starting from a cold-start recall of 0.8877, Neural Sparse converges to a higher plateau of 0.9627 versus 0.9548 for Trigram Fuzzy, recovering 7.5\% more valid entities relative to the cold-start state and delivering an absolute gain of +0.79\% over the production baseline. These results demonstrate the model's superior capability in surfacing long-tail candidates within the production environment, thereby effectively enhancing the overall HCI learning rate.

\begin{table}[h]
\caption{Simulation results over 7 days of behavioral logs. Neural Sparse achieves better HCI coverage. Recall measured at k=25 to match production settings.}
\label{tab:simulation}
\small
\begin{tabular}{lcc}
\toprule
\textbf{Configuration} & \textbf{Recall@25} & \textbf{vs. Cold Start} \\
\midrule
Cold start (no HCI) & 0.8877 & --- \\
HCI Exact + Trigram Fuzzy & 0.9548 & +6.7\% \\
HCI Exact + Neural Sparse & \textbf{0.9627} & +7.5\% \\
\midrule
\multicolumn{3}{l}{\textit{Absolute improvement: +0.79\% over Trigram Fuzzy}} \\
\bottomrule
\end{tabular}
\end{table}

\section{Conclusion}

We presented a neural sparse retrieval system for music search that significantly outperforms traditional n-gram matching on fuzzy query matching tasks. By training a custom granular tokenizer (max 3-char tokens) combined with a domain-specific BERT model on music search queries, we achieve an aggregate \textbf{91.4\%} recall@10 at production-comparable throughput. End-to-end simulation shows consistently improved exploration efficiency over trigram matching. 
Our ablation studies demonstrate that the sparse training methodology is the primary driver of performance, and that a modest in-domain corpus is a cost-effective substitute for large-scale general-purpose pretraining. Our system demonstrates the viability of granular-tokenizer neural sparse methods for large-scale commercial search applications where latency and integration constraints are critical.

\begin{acks}
We thank the Amazon Music Search and Amazon OpenSearch teams for their support and infrastructure.
\end{acks}

\section*{Speaker Biography}
Paul Greyson is a Principal Engineer for search at Amazon Music where he currently leads development of machine learning based search solutions. He has worked in music technology for more than 25 years and received a bachelor's degree in Computer Science from UC Berkeley.

\bibliographystyle{ACM-Reference-Format}
\bibliography{sample-base}

\end{document}